# Flow Guided Short-term Trackers with Cascade Detection for Long-term Tracking

Han Wu    Xueyuan Yang    Yong Yang    Guizhong Liu

School of Information and Communication Engineering, Xi'an Jiaotong University

Xianning West Road 28, 710049, Xi'an, P.R. China

xjtuwh@stu.xjtu.edu.cn    yxy1995@stu.xjtu.edu.cn    yy1996@stu.xjtu.edu.cn    liugz@xjtu.edu.cn

**Abstract**

*Object tracking has been studied for decades, but most of the existing works are focused on the short-term tracking. For a long sequence, the object is often fully occluded or out of view for a long time, and existing short-term object tracking algorithms often lose the target, and it is difficult to re-catch the target even if it reappears again. In this paper a novel long-term object tracking algorithm flow_MDNet_RPN is proposed, in which a tracking result judgement module and a detection module are added to the short-term object tracking algorithm. Experiments show that the proposed long-term tracking algorithm is effective to the problem of target disappearance.*

## 1. Introduction

Single object tracking (SOT) is the problem of estimating the trajectory of a target in a sequence of images [1]. It is challenging since only the initial state of the target is known. The ability to track an arbitrary object would be useful for many applications including video analytics, surveillance, robotics, augmented reality and video editing. The requirement to track anything given only a single example presents a significant challenge due to the many complex factors that affect the appearance, including aspect ratio change, background clutter, camera motion, fast motion, full occlusion, illumination variation, low resolution, out of view, partial occlusion, similar object, scale variation and viewpoint change. In particular, in a long video, the object may leave the field of view or be fully occluded for a long period and reappear again. Traditional tracking algorithms have focused on the problem of short-term tracking which does not require methods to perform re-detection [2, 3, 4]. This implies that the object is always present in the video. However, Long-term tracking is more practical for realistic systems since the target is easily out of view in long video. For most practical applications, it is critical to track objects through disappearance and re-appearance events, and further, to be aware of the presence or absence of the object. The long-term tracking does not just refer to the sequence length, but more importantly to the sequence properties (number of target disappearances and reappears, etc.) [5]. The tracker should report the target position in each frame when the target presents and provide a confidence score of the target presence.

Only a few datasets have been proposed in long-term tracking. The first dataset is introduced by the LTDT challenge (http://www.micc.unifi.it/LTDT2014/), which offers a collection of specific videos. And [6] proposes the UAV20L dataset for low-altitude UAV target tracking and it contains twenty long sequences with many target disappearances recorded from drones. [5, 7, 8] propose three benchmarks that datasets contains many target disappearances. VOT2018 introduces a long-term tracking sub-challenge to the set of standard VOT sub-challenges and it consists of 35 long sequences formed by 146, 847 frames containing many target disappearances [9], and the shortest video includes 1, 389 frames and the longest includes 29, 700 frames. The Vision Meets Drone Single-Object Tracking (VisDrone-SOT2019) challenge collects 132 video sequences divided into three non-overlapping sets, i.e., training set (86 sequences with 69, 941 frames), validation set (11 sequences with 7, 046 frames), and testing set (60 sequences with 112, 011 frames). It contains long sequences as well as short sequences [10].

The goal of this paper is to increase the robustness of object tracking by close cooperation between failure judgement and detection module based on short-term tracking. The main contributions of our work are summarized below.

i. We propose a long-term tracking framework, which contains short-term tracking module, judgement module and detection module.
ii. The proposed judgement module can accurately judge whether the target is present or not, that is, whether the target is fully occluded or out of view.
iii. The proposed detection module can capture the target as soon as the target reappears.

The rest of the paper is organized as follows. We first review the related works in Section 2, and present our *flow_MDNet_RPN* approach for long-term tracking in Section3. Section4 demonstrates the experimental results. Finally, we summarize our work in Section 1.

## 2. Related Work

The research of single object tracking problem has been a long time. Many related works have thoroughly studied about it.

### 2.1. Short-term trackers

Most state-of-the-art approaches follow the tracking-by-detection paradigm, where a classifier or regressor is discriminatively trained to differentiate the target from the background [1]. The current tracking algorithms are mainly divided into two categories: some are based on correlation filter, and the others are based on deep learning.

**Based on correlation filter.** Among tracking-by-detection approaches, the Discriminative Correlation Filter (DCF) based trackers have recently shown excellent performance on the standard short-term tracking benchmarks [11], [12]. The key for their success is the ability to efficiently utilize limited data by including all shifts of local training samples in the learning. DCF-based methods train a least-squares regressor to predict the target confidence scores by utilizing the properties of circular correlation and the Fast Fourier Transform (FFT) [1]. The seminal work that puts forward correlation filter to tracking is MOSSE, which uses a set of samples random affine transformed from the initial target to construct a minimum output sum of squared error filter [13]. The correlation filter transforms the object template matching problem into a correlation operation in the frequency domain. A series of excellent short-term trackers based on correlation are proposed such as KCF [2], DSST [14], SRDCF [15] and ECO [3].

**Based on deep learning.** With the development of deep learning in recent years, more and more researchers apply deep learning methods to object tracking to improve the tracking effect. [4] proposes a multi-domain learning framework (MDNet) based on CNNs, which separates domain-independent information from domain-specific one, and the CNN pretrained by multi-domain learning is updated online in the context of a new sequence to learn domain-specific information adaptively. MDNet gets outstanding performance on several short-term tracking benchmarks. Fully-convolutional Siamese network is another attractive way due to its state-of-the-art performance as well as high efficiency. The representative trackers include SiamFC [16], GOTURN [17], CFNet [18] and SINT [19]. These Siamese trackers formulate the visual object tracking problem as learning a general similarity map by cross-correlation between the feature representations learned for the target template and the search region. The network consists of Siamese deep neural network and correlation layers are trained end-to-end off-line with large-scale image pairs. SiamRPN [20], DaSiamRPN [21] and SiamRPN++ [22] formulate the tracking as a one-shot local detection task by introducing a region proposal network following a Siamese network and they get top performance with real-time speed. These trackers simply match the initial patch of the target in the first frame with candidates in a new frame and return the most similar patch by a learned matching function [19].

### 2.2. Long-term trackers

Long-term tracking has received far less attention than short-term tracking. Superior short-term trackers have poor performance on very long sequences due to the localization errors accumulation and the updates gradually deteriorate their visual model, leading to drift and failure. To avoid tracker drift, LCT [26], ECO [3] and MDNet [4] adopt a conservative updating mechanism. A major difference between long-term and short-term tracking is that long-term trackers are required to handle situations in which the target may leave the field of view for a long duration. Once tracking failed, the short-term trackers usually cannot identify the false. Failure recovery, however, is primarily addressed in long-term trackers [5]. It means that long-term trackers have to detect target absence and re-detect the target when it reappears. A typical structure of a long-term tracker is a short-term component with a relatively small search range responsible for frame-to-frame association and a detector component responsible for detecting target reappearance, e.g. TLD, which tracks by median flow and detects by random fern classifier [25]. In addition, an interaction mechanism between the short-term component and the detector is required that appropriately updates the visual models and switches between target tracking and detection [25]. The tracker should output the confidence of the tracking result and judge the object is present or not.

Many short-term trackers can report the tracking confidence by using their visual model similarity scores at the reported target position, but they can't re-detect the target once the target reappears far from the previous position. LCT trains an online classifier to re-detect objects in case of tracking failure, however, the proposed failure criterion is suspect and the tracker just detects object around the tracking result. DaSiamRPN_LT [21] performs long-term tracking by introducing a local-to-global search region strategy, which improves the performance of the tracker in out-of-view and full occlusion challenges. The winner of VisDrone-SOT2018, LZZ-ECO [23], utilizes deep object detector YOLOv3 [24] to detect the target. MBMD, the winner of VOT2018 long-term tracking challenge, proposes a long-term tracking framework based on deep regression and verification network [27]. Similar to MBMD, we propose a long-term tracker which can report the target position along with a presence confidence score, and the tracker can detect the target when the object reappears. Different from MBMD, we use a better short-term tracking method. And the failure determination mechanism and detection method are novel.

## 3. Proposed Long-term Tracking Approach

This section is organized as follows. Section 3.1 is the overview of the proposed long-term tracking approach. The short-term tracking module is detailed in Section 3.2. The judgement module is described in Section 3.3 and the detection module is described in Section 3.4. The flow chart of proposed method is presented in Section 3.5.

### 3.1. Overview

To continually tracking the target in a video even it may be occluded or out of view for a long time, we propose a novel long-term tracker which combines three components, a short-term tracking module, a judgement module and a cascade detection module together. The tracker is initialized based on the initial target state. Due to the inter-frame association, the short-term tracking module only needs to search locally around target position in previous frame. Once the short-term tracking module reports a tracking result, the judgement module judges the target is present or absent and outputs the tracking box's confidence. If the target disappears, or short-term tracking failed, the detection module performs a cascade detection around last tracking box from local to global to capture the target once it reappears. The overall procedure is shown as Figure 1.

### 3.2. Short-term tracking module

MDNet [4] is composed of shared layers and multiple branches of domain-specific layers, where domains correspond to individual training sequences and each branch is responsible for binary classification to identify the target in each domain. At the first frame, MDNet is initialized by the specific target and online tracking at next frame. The new classification layer and the fully connected layers within the shared layers are then fine-tuned online during tracking to adapt to the new domain. Online tracking by MDNet is performed by evaluating the candidates randomly sampled around the previous target state which obey a gaussian distribution.

SiamRPN++ [22] obtains state-of-the-art results on the VOT2018 [9] short-term tracking in real time. SiamRPN++ is composed of a multi-layer aggregation module which assembles the hierarchy of connections to aggregate different levels of representation and a depth-wise correlation layer which allows the network to reduce computation cost and redundant parameters while also leading to better convergence. The network is pretrained offline and it does not update online which avoids tracking model drift.

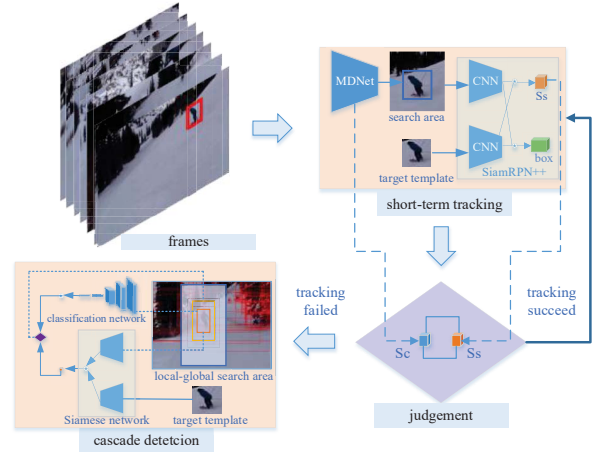

Figure 1. The overall framework of the proposed long-term tracking approach which contains a short-term tracking module, a judgement module and a cascade detection module.

We find that the MDNet and SiamRPN++ can complement with each other. MDNet's online updating mechanism makes it is able to adapt to target changes and it has stronger discriminating ability. Therefore, we combine the two trackers as our short-term tracking module for joint tracking. MDNet outputs tracking candidate object. SiamRPN++ fine-tunes the candidate and outputs a similarity score $S_s$ to measure the similarity between the candidate object and the target. The MDNet online updates to adapt to the object appearance variations based on newly reliable observations. If tracking confidence is higher than the designed threshold $th_{mid} = 0.5$, the multi-domain network samples positive and negative samples from current frame and when tracking confidence is lower than the fixed threshold $th_{low} = 0.1$, the network updates the classifier based on the sample pairs to adapt to the changes in the appearance of the target.

### 3.3. Judgement module

The short-term tracking module generates tracking box, but it can't identify object is present or absent. A judgement module that is responsible for estimating tracking box is introduced. We utilize multi-domain network and Siamese network to verify the tracking result of short-term module. When the short-term tracking box comes on each round, the multi-domain network outputs its classification score $S_c$ and the Siamese network predicts its similarity score $S_s$. Next, the algorithm makes decisions based on $S_s$ and $S_c$. The decision-making process is shown in Figure 2.

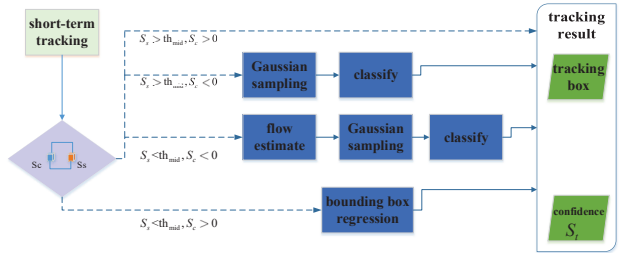

Figure 2. We make different decisions according to $S_s$ and $S_c$. The solid line indicates that there must be a connection. The dotted line indicates that there is not necessarily a connection.

If $S_s > th_{mid}(th_{mid} = 0.5)$ and $S_c > 0$, the tracking is successful. If $S_s > th_{mid}$ and $S_c < 0$, the tracker may meet distractors, and then Gaussian sampling around the tracking results and the candidate target boxes with the highest classification score is selected. The Gaussian sampling means the candidates randomly sampled around the previous target state which obey a gaussian distribution. If $S_s < th_{mid}$ and $S_c < 0$, the tracker may follow the background, and the optical flow between previous frame and current frame is extracted to guide Gaussian sampling. PWC-Net [32] extracts flow information and gets motion vector field of pre-frame and current frame. We add the motion vector of the object to the target position in previous frame to compensate global motion on the pre-frame object state. The corresponding target box with the highest classification score is selected. If $S_s < th_{mid}$ and $S_c > 0$, the current tracking result box is inaccurate, and the bounding box regression fine-tunes current tracking candidate box. Follow the practice of MDNet, the bounding box regression is a simple linear regression model to predict the precise target location using conv3 features of the samples near the target location. The bounding box regressor is trained only in the first frame since it is time consuming for online update and incremental learning of the regression model may not be very helpful considering its risk. Refer to [4] for details as we use the same formulation and parameters. The confidence $S_t$ which is calculated by the Siamese network is the similarity of current tracking box between the initial target template. If the confidence of current tracking box is less than $th_{low}$, we consider the short-term tracking failed and start the detection module.

### 3.4. Cascade detection module

When current tracking is failed, the algorithm switches to detection module. It is critical to expand the search region to ensure that the target is able to be detected by the tracker. We propose a cascade detection method to capture the target in a very short time once the target reappears. The main module is shown in Figure 3.

When the target is out of view or fully occluded, the same as Section 3.3, the optical flow network PWC-Net [32] estimates the global motion of two adjacent frames. Gaussian sampling is then performed around the target position of the previous frame, the multi-domain network classifies the sampled candidates and returns the tracking box with the highest classification score. The similarity score between the initial target template is calculated by the Siamese network. If both the classification score and the similarity score are greater than the set threshold, the target is considered to be found, otherwise the search area is expanded. We sample a rectangle patch centered at the target, with an area of about $5^2$ times the target area. Since the target is not required to belong to any set of pre-defined classes or be represented in any existing training datasets, a novel region proposal network, named GA-RPN [33], which leverages semantic features to guide the anchoring is used to generate sparse candidate boxes. By predicting the scales and aspect ratios instead of fixing them based on a predefined list, the scheme handles tall or wide objects more effectively. The candidate box with the highest classification score and similarity score is selected. If both the classification score and the similarity score are greater than the set threshold, the target is considered to be found, otherwise the search area will be expanded again and we sample a rectangle patch centered at the target, with an area of about $18^2$ times the target area, and candidate target boxes are generated using GA-RPN. The target box with the highest classification score and similarity score is selected. If both the classification score and the similarity score are greater than the set threshold, the target is considered to be found, otherwise the search area will be expanded to the global frame. GA-RPN generates candidate target boxes. The best proposal is selected step by step according to the position, distance, shape and appearance information. If both the classification score and the similarity score of the candidate box are greater than the set threshold, the target is considered to be found, otherwise, we continue to detect target at the next frame. If the target is found, the algorithm switches to short-term tracking module.

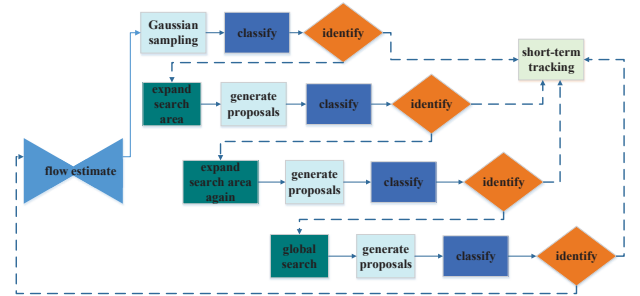

Figure 3. The framework of proposed cascade detection module, which expand search area from local to global frame.

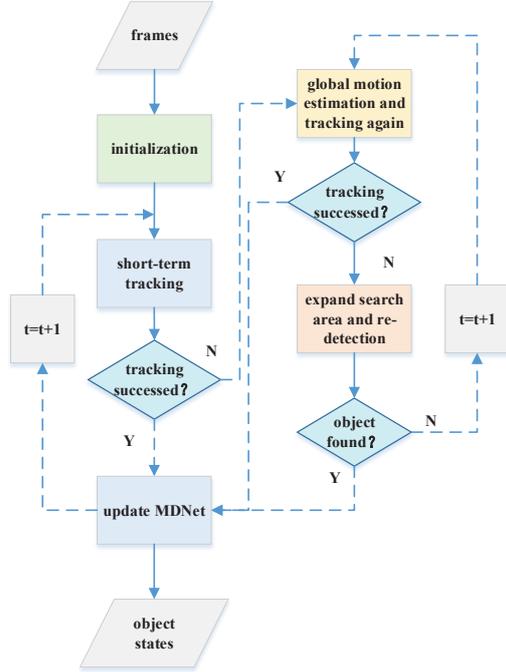

Figure 4. Flow chart of proposed *flow_MDNet_RPN*.

### 3.5. Flow chart of the proposed method

The tracking steps of the *flow_MDNet_RPN* is shown in the Figure 4. The detailed steps are as follows.
**Formal description of the algorithm *flow_MDNet_RPN***
i. Input the frames of a sequence.
ii. Train MDNet fully connected layers by initial state and complete initialization for other modules.
iii. Start short-term tracking to get a tracking bounding box.
iv. Judge tracking succeed or failed by MDNet classification score and Siamese similarity score. If tracking succeed, the algorithm continues short-term tracking.
v. If tracking failed, start cascade detection. And judge tracking succeed or failed again.
vi. Output estimated target states.

## 4. Experiments

We perform a lot of experiments and evaluate the performance of the proposed *flow_MDNet_RPN*.

### 4.1. Implementation Details

The architecture of the MDNet receives a 107 x 107 RGB input, and has five hidden layers including three convolutional layers (conv1-3) and two fully connected layers (fc4-fc5). The MDNet use VGG-M [35] to extract general object features and the network is pretrained on ImageNet VID dataset [28].

| Tracker | F-score | Pr | Re |
|---|---|---|---|
| SiamRPN++ | 0.5069 | **0.6766** | 0.4053 |
| MDNet | 0.3866 | 0.3732 | 0.4010 |
| Ours | **0.5405** | 0.6095 | **0.4856** |

Table 1. Performance evaluation for algorithms on the VOT-2018 LTB35 dataset. The best result is marked in bold.

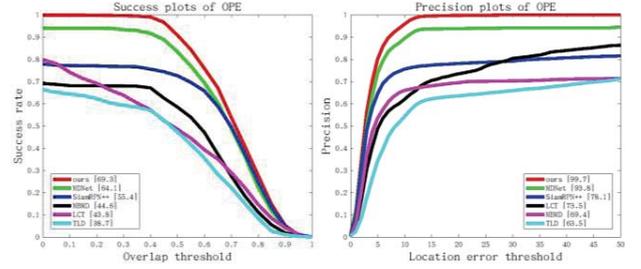

Figure 5. Success plot and precision plot of VisDrone-SOT2019 validation dataset. Our method is shown by red curves.

The SiamRPN++ model is pretrained on COCO dataset [34], ImageNet DET [28], ImageNet VID [28] and Youtube-BoundingBoxes dataset [29] and the backbone is ResNet-50 [30]. The two streams of the network share the same architecture but the sizes of the inputs are different. The size of search area image patch is 255 x 255 x 3, the size of target templates is 127 x 127 x 3. There are 4 anchors, so the network output a 25 x 25 x 2 x 4 tensor for classification and a 25 x 25 x 4 x 4 tensor for regression. The anchor ratios are [0.33, 0.5, 1, 2, 3].

We use PWC-Net trained on FlyingChairs [31] and FlyingThings3D [36] datasets to extract optical flow as our motion estimation network.

Our proposed tracking algorithm is implemented in PyTorch with 3.50GHz Intel Xeon(R) CPU E5-1620 and NVIDIA GTX1080Ti GPU.

### 4.2. State-of-the-art Comparisons

**VOT-2018 LTB35 dataset.** We use VOT2018-LT [9] to evaluate the proposed tracker. The detailed comparisons are reported in Table 1. From Table1, we can conclude that the proposed tracker achieves state-of-the-art performance in terms of **F-score**, **Pr** and **Re** criteria. The tracking precision (**Pr**), recall (**Re**) and F-score metrics are utilized for accuracy evaluation. The threshold F-measure $F(\tau_\theta)$ is defined as

$$F(\tau_\theta) = 2\Pr(\tau_\theta)\text{Re}(\tau_\theta)/(\Pr(\tau_\theta)+\text{Re}(\tau_\theta)). \quad (1)$$

Where $\tau_\theta$ is the threshold. Then, the F-score is defined as the highest score on the F-measure plot over all thresholds.

From Table1. Our algorithm gets highest **F-score** 0.5405 and **Re** score 0.4856. Notice that the **Pr** score 0.6095 is lower than SiamRPN++. This is that detection results may be wrong, which directly affects subsequent tracking results.

This means that exploring better re-detection algorithms to accurately identify the specific target is crucial for tracking robustness. And in the long-term tracking, the detection module is critical to the challenge of the target out of view.

**VisDrone-SOT2019 validation dataset.** There are 11 sequences in VisDrone-SOT2019 validation dataset. Following the evaluation methodology in [10], we use the success and precision scores to evaluate the performance of the proposed tracker. The success score is defined as the area under the success plot. That is, with each bounding box overlap threshold $t_o$ in the interval [0, 1], we compute the percentage of success fully tracked frames to generate the successfully tracked frames vs. bounding box overlap threshold plot. The overlap between the tracker prediction $B_t$ and the groundtruth bounding box $B_g$ is defined as $O = |B_t \cap B_g|/|B_t \cup B_g|$ where $\cap$ and $\cup$ represent the intersection and union between the two regions, respectively, and $|\cdot|$ calculates the number of pixels in the region. Meanwhile, the precision score is defined as the percentage of frames whose estimated location is within the given threshold distance of the ground truth based on the Euclidean distance in the image plane. Here, we set the distance threshold to 20 pixels in evaluation for comparison.

In this experiment, we compare our method with several representative trackers, including MDNet [4], SiamRPN++ [22], MBMD [27], TLD [25] and LCT [26]. As shown in Figure 5, ours is much higher than the performance of the baseline algorithm MDNet and SiamRPN++. Our method gets 69.3 success scores and 99.7 precision scores which improves 5.2 and 5.9 higher success and precision scores compared with MDNet and it improves 13.9 and 21.6 higher success and precision scores compared with SiamRPN++. This is due to the introduction of the re-detection module. The success scores and precision scores of our algorithm are much higher than the existing classic long-term object tracking algorithm TLD and LCT. This is due to the powerful discriminative classification ability of the convolutional neural network. Our algorithm is even better than the MBMD equipped with the re-detection module. We analyze that it's because the failure detection module use ad-hoc rules to make decision and the threshold of failure judgement which is pre-defined related to the scene closely, and our proposed algorithm which has been trained on the VisDrone-SOT2019 training dataset is more suitable for video taken by drones. Further work can investigate the adaptive failure judgement mechanism for different scenes.

**VisDrone-SOT2019 testing dataset.** There are 60 sequences in VisDrone-SOT2019 testing dataset. Evaluating on more data can more accurately reflects the robustness of the tracking algorithm. Note that the number of frames in the testing dataset (60 sequences with 112, 011 frames) is more than the sum of the training dataset (86 sequences with 69, 941 frames) and the validation dataset (11 sequences with 7, 046 frames). It is because that the testing dataset contains a lot of long videos in which the target may be out of view or fully occluded for a long time.

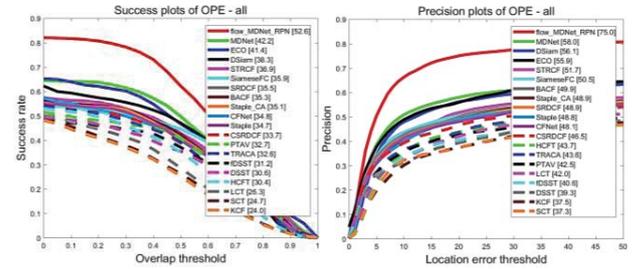

Figure 6. Success plot and precision plot of VisDrone-SOT2019 testing dataset. The proposed method *flow_MDNet_RPN* is shown by red curves.

As shown in Figure 6, our proposed *flow_MDNet_RPN* gets 52.6 success scores and 75 precision scores which improves 10.4 and 17 higher success and precision scores compared with our baseline MDNet. It demonstrates the outstanding performance of our tracking algorithm compared to the state-of-the-art techniques such as ECO [3]. The performance boosts not only rely on the robust visual models but also the capability to detect object when the object has been out-of-view for a long time and reappears in the view.

From the performance of our tracker in terms of different attributes in Table 2 we find that Low Resolution (LR) performance scores are the lowest which means that further research can be done on how to improve the tracking robustness for weak small targets. Similar Object (SOB) is another challenge that damages tracking performance. More results analysis can refer to D. Du, P. Zhu, L. Wen, X. Bian, H. Ling, Q. Hu, and et al. VisDrone-SOT2019: The Vision Meets Drone Single Object Tracking Challenge Results. 2019.

**Visual Results.** To visualize the performance of our tracker, we provide representative results of our tracker and the other two baseline methods. Video frames are derived from VisDrone-SOT2019 dataset. As shown in Figure 7, at the beginning, all three algorithms can track the object well. However, SiamRPN++ often tracks distractors with the same semantic class. MDNet is able to track object stably, but it cannot re-capture the target when the tracking failed. Our algorithm not only perform short-term tracking stably, but also can detect the target after short-term tracking failed. Through robust short-term tracking, judgement mechanism and the re-detection module, the tracker complete the localization of the specific target in the video. The short-term object tracking equipped with the detection module has a certain improvement.

| Attributes | ARC | BC | CM | FM | FOC | IV | LR | OV | POC | SOB | SV | VC | all |
|---|---|---|---|---|---|---|---|---|---|---|---|---|---|
| Precision | 74.7 | 77.6 | 76.8 | 73.6 | 63.2 | 76.1 | 69.9 | 66.3 | 65.2 | 71.6 | 84.4 | 81.4 | 75.0 |
| Success | 57.6 | 52.4 | 55.4 | 54.9 | 44.4 | 53.4 | 42.1 | 54.1 | 46.2 | 44.3 | 62.4 | 58.2 | 52.6 |

Table 2. Success and precision scores of our method on VisDrone-SOT2019 testing dataset in terms of 12 attributes including Aspect Ratio Change (ARC), Background Clutter (BC), Camera Motion (CM), Fast Motion (FM), Full Occlusion (FOC), Illumination Variant (IV), Low Resolution (LR), Out of View (OV), Partial Occlusion (POC), Similar Object (SOB), Scale Variation (SV) and Viewpoint Change (VC).

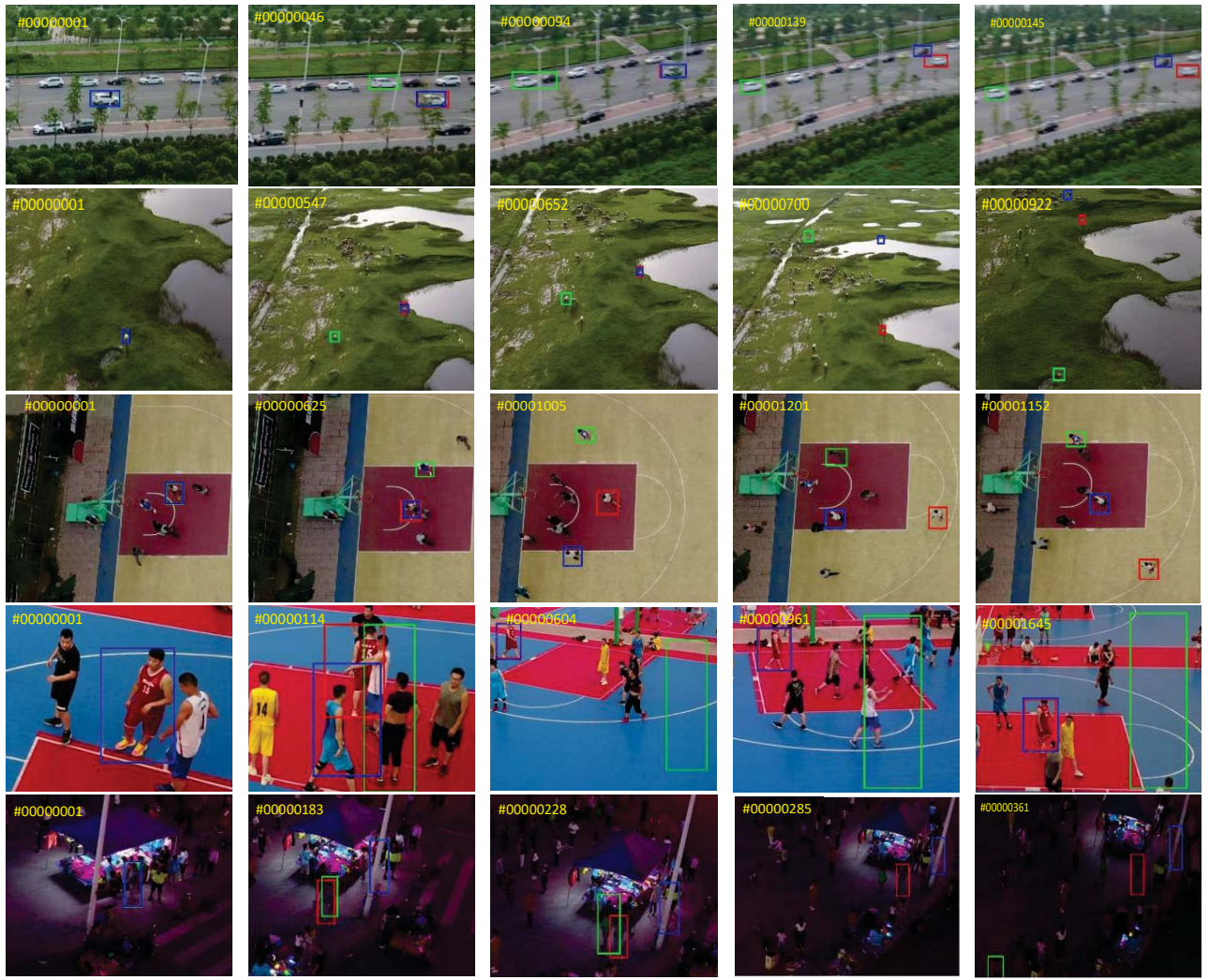

Figure 7. Visual results of our tracker, along with SiamRPN++ [21] and MDNet [4]. The red box represents ours, the green box represents SiamRPN++ and the blue box represents MDNet. Video frames are derived from VisDrone-SOT2019 dataset. From top to bottom: uav0000003_00000, uav0000092_01150, uav0000080_01680, uav0000085_00000, uav0000071_02520.

## 5. Conclusion

In this paper, a long-term object tracking scheme *flow_MDNet_RPN* is proposed, which mainly consists in the addition of a judgement module and a cascade detection module to a short-term object tracking algorithm. The short-term tracking algorithm integrates MDNet and SiamRPN++. The judgement module combines classification score and similarity score. The cascade detection module is guided by optical flow. Experiments show that the proposed long-term tracking algorithm is effective to the problem of target disappearance.